\documentclass[10pt,twocolumn,letterpaper]{article}

\usepackage{cvpr}
\usepackage{times}
\usepackage{epsfig}
\usepackage{graphicx}
\usepackage{amsmath}
\usepackage{amssymb}

\usepackage[breaklinks=true,bookmarks=false]{hyperref}

\cvprfinalcopy % *** Uncomment this line for the final submission

\def\cvprPaperID{****} % *** Enter the CVPR Paper ID here

\begin{document}

%%%%%%%%% TITLE
\title{Can Adversarial Networks Hallucinate Occluded People With a Plausible Aspect?}

\author{
    Federico Fulgeri \quad \quad Matteo Fabbri \quad \quad Stefano Alletto \quad \quad Simone Calderara \quad \quad Rita Cucchiara \\
    \\
    University of Modena and Reggio Emilia\\
    Via P. Vivarelli 10, Modena, 41125, Italy
}

\maketitle
%\thispagestyle{empty}

%%%%%%%%% ABSTRACT
\begin{abstract}
When you see a person in a crowd, occluded by other persons, you miss visual information that can be used to recognize, re-identify or simply classify him or her. You can imagine its appearance given your experience, nothing more. Similarly, AI solutions can try to hallucinate missing information with specific deep learning architectures, suitably trained with people with and without occlusions.
The goal of this work is to generate a complete image of a person, given an occluded version in input, that should be a) without occlusion b) similar at pixel level to a completely visible people shape c) capable to conserve similar visual attributes (e.g. male/female) of the original one.
For the purpose, we propose a new approach by integrating the state-of-the-art of neural network architectures, namely U-nets and GANs, as well as discriminative attribute classification nets, with an architecture specifically designed to de-occlude people shapes. The network is trained to optimize a Loss function which could take into account the aforementioned objectives. As well we propose two datasets for testing our solution: the first one, occluded RAP, created automatically by occluding real shapes of the RAP dataset created by \cite{rap} (which collects also attributes of the people aspect); the second is a large synthetic dataset, AiC, generated in computer graphics with data extracted from the GTA video game, that contains 3D data of occluded objects by construction. Results are impressive and outperform any other previous proposal. This result could be an initial step to many further researches to recognize people and their behavior in an open crowded world.
\end{abstract}

%%%%%%%%% BODY TEXT
\section{Introduction}
\label{sec:intro}

While recent efforts in people detection, recognition, and tracking enabled a plethora of video-surveillance applications, e.g. people identification by \cite{reid}, pose estimation by \cite{Guler2018DensePose} and action analysis by \cite{action}, occlusion is still an open problem.
The occlusion issue is well known in the people detection and tracking literature and generally affects any intelligent video surveillance system, but it is debatable whether a real solution to the problem could exist effectively.
In fact, whenever an occlusion occurs we observe a removal of information from the observed scene. The occluded portion of an object, indeed, becomes unknown and, in a Parmenidean sense, it does not exist until it can be observed. For this motivation, most of the literature focused on counteracting the phenomenon conveying occlusion robustness to either detection, tracking, or re-id systems as by \cite{reid1,reid3,track1,track2,track3}. 
In the matter of fact, recovering the image content from an occlusion is feasible only in the case where the target has been previously observed e.g. in a video stream. This is the approach followed also by many tracking solutions which memorize several detected appearance of the person, to discard occlusions as ``less frequent accidents'' w.r.t. the normal visible appearance. Nevertheless, leveraging on the generative capabilities of GANs by \cite{godgan}, we aim at demonstrating that it is indeed possible to hallucinate a plausible representation of the occluded content even when it has never been previously observed, i.e. in single images.

\begin{figure*}[h!]
\centering
\includegraphics[width=0.95\textwidth]{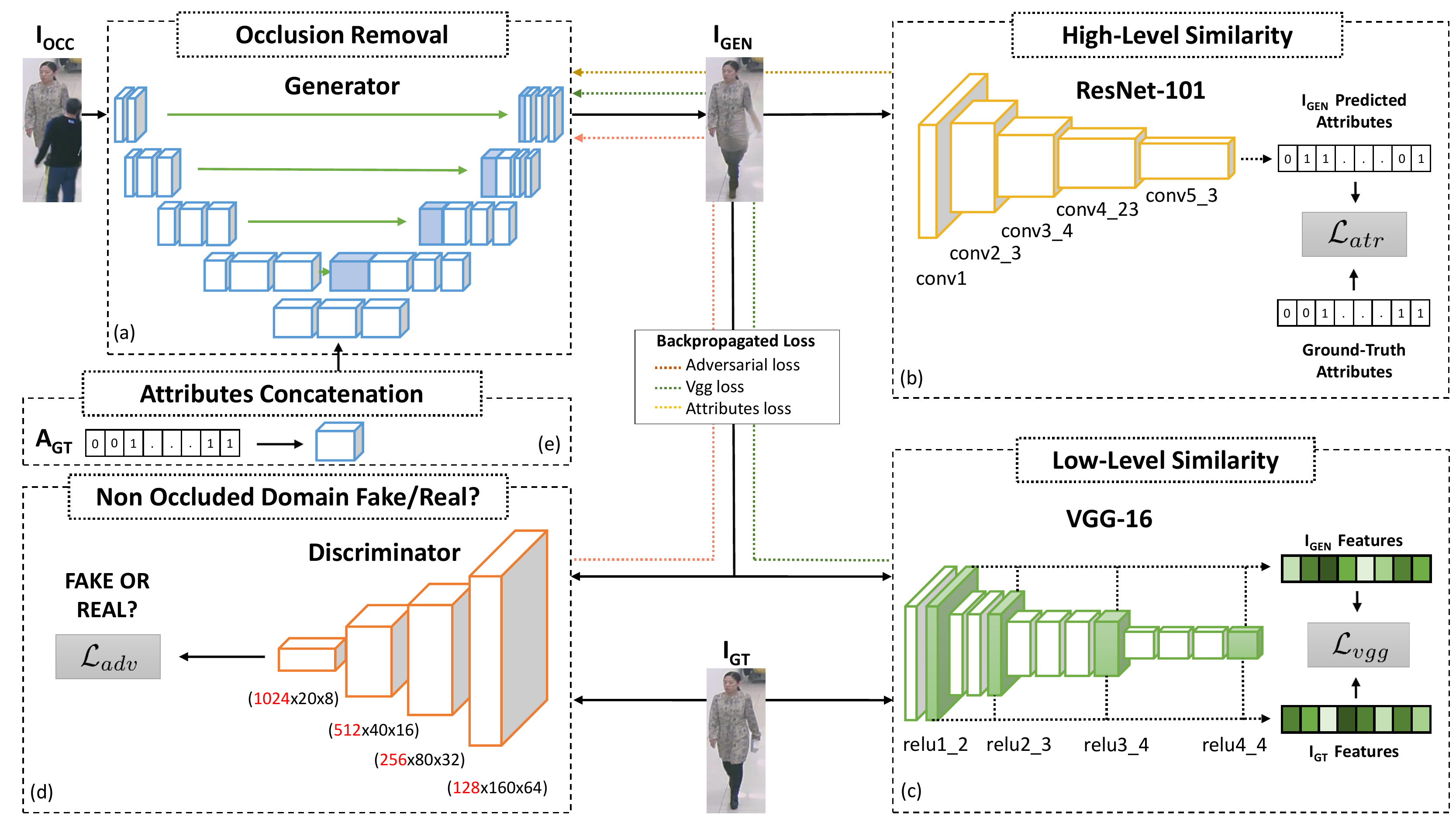}
\caption{A schematic representation of the training procedure adopted in our work. The Generator (a) takes the occluded image $I_{OCC}$ as input and the attributes of the person $A_{GT}$ (e) as a further conditioning element. The output of the Generator $I_{GEN}$ is the restored image, with no occlusion. To train the Generator, we fed $I_{GEN}$ to three different networks: ResNet-101, VGG-16, and the Discriminator. (b) ResNet-101 gives a prediction of the $I_{GEN}$ attributes which are compared with the ground-truth ones for the $\mathcal{L}_{atr}$ computation, in order to maximize high-level similarity. (c) The feature maps extracted from different layers of VGG-16 are used to calculate the content loss between $I_{GEN}$ and $I_{GT}$ with the aim of encouraging low-level similarity. (d) The Discriminator, which gives the judgment between ``real'' and ``fake'' distributions, has to be fooled by the Generator in order to produce images belonging to the non-occluded domain of pedestrians. The Discriminator is trained alongside the Generator to distinguish between generated ``fake'' images and ``real'' ones. At evaluation time, only the Generator network and the Attributes Concatenation are used.}
\label{fig:complete_net}
\end{figure*}
Following on our previous work on the topic (\cite{fabbri2017generative}), in this paper, we introduce a novel network that leverages the generative power of GANs for hallucinating the occluded portion of the image without any guidance of an attention mechanism that could provide instance level information about the occluding person. The proposed solution aims at generating or reconstructing the image of a person which could be plausible in many senses: a) similar to images of real people, observed in the training dataset; b) acceptable at pixel level as a person shape; c) capable to preserve similar visual attributes of the ground truth de-occluded image. This is carried out by exploiting solutions for attribute classifications (e.g. male/female, young/old, with/without trousers, etc.) and integrating them in a U-net like generative and adversarial architecture. 

Another major problem that arises when dealing with occlusions, through learning-based solutions, is the lack of large-scale datasets providing realisticly occluded and non-occluded pairs of images. 
Most of the proposed solution in literature, like the ones introduced by \cite{fabbri2017generative,ouyang2016partial,op2015detection} paste together different people detections, or manually add random objects or textures to a non-occluded image. These processes ultimately fail to generate realistic data and are thus a liability when employed for training a neural network that aims at resolving the occlusion while keeping the rest of the image coherent (e.g. the background) and preserving the person's attributes.
To address this issue, we propose a novel, fully automatic, way to generate realistic occlusion pairs by exploiting the recent achievements in object segmentation by \cite{he2017mask}. These results are high-fidelity occlusion pairs, where the background of the original image is preserved and the generated occlusion is more realistic. 
Additionally, thanks to the software provided by \cite{fabbri2018learning}, we created the massive computer graphics generated AiC dataset (leveraging on the highly photo-realistic graphics of GTA V video-game), in which we artificially created a large collection of occluded pedestrians. Additionally, we recovered from the game engine their attributes by manually annotating just the models. To our knowledge, this is the first CG dataset for the purpose of de-occluding people having a set of annotated person attributes (e.g. sex, hair color, dress style, etc.). The dataset is pubblicly available \footnote{http://aimagelab.ing.unimore.it/aic}.

To summarize, our contributions are threefold:
\begin{itemize}
\item We propose a novel generative adversarial network that is able to solve occlusions in pedestrian images by hallucinating the missing parts while keeping both the appearance and the background coherent;
\item We devise a new way for synthetically generating occlusion pairs that result in more realistic images when compared to other methods previously employed, also by creating a huge CG dataset;
\item We propose a method for conditioning the occluded body part restoration on pedestrian attributes and consequently improving the generation process.
\end{itemize}
We show by experiments that the design of a conditional GAN that is aware of the attributes can acceptably hallucinate pedestrian and we experimentally demonstrate that this information is helpful in guiding the generation process.
Results are interesting in terms of very high accuracy, outperforming other previous methods. We believe that our method could be useful in many computer vision systems, from surveillance, automotive to human behavior understanding tasks.

\begin{figure}[ht!]
\centering
\includegraphics[width=0.48\textwidth]{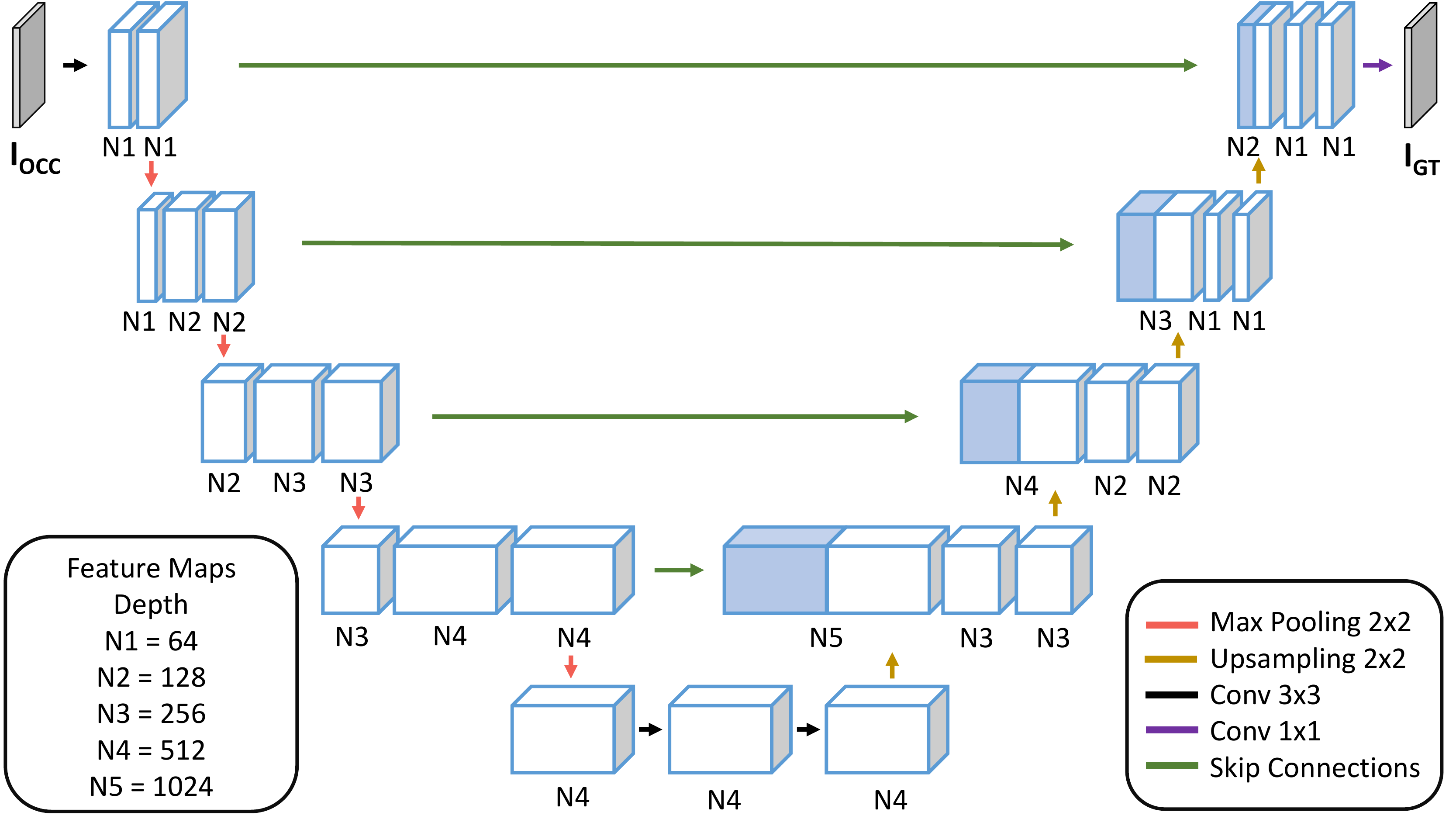}
\caption{Architecture of our Generator network with corresponding number of feature maps and kernel sizes. The figure also depicts max-pooling and upsampling operations, along with skip connection gates.}
\label{fig:genarch}
\end{figure}

\section{Related Works}
Generative image modeling with deep learning techniques has received lots of attention in recent years. Works on this field can be split into two categories. The first line of works follows the unsupervised setup. Here, the variational autoencoders (VAE) proposed by \cite{rezende2014stochastic} and \cite{kingma2013auto} are the first popular methods which apply a re-parameterization trick to maximize the lower bound of the data likelihood. The most popular methods are indeed generative adversarial networks (GAN) of \cite{godgan} and \cite{radford2015unsupervised}, which simultaneously learn a generator network to generate image samples, and a discriminator network to discriminate generated samples from real ones. GANs are capable of generating sharp images by exploiting the adversarial loss instead of more canonical losses such as L1 or L2.

The second group of works produces images conditioned on either categories, attributes, labels, images or texts. \cite{yan2016attribute2image} proposed a Conditional Variational Autoencoder (CVAE) to achieve an image generation conditioned on attributes. On the other hand, \cite{mirza2014conditional} proposed conditional GANs (CGAN) where both the generator and the discriminator are conditioned on extra information to perform category specific image generation. \cite{lassner2017generative} generated people in clothing, by conditioning on the fine-grained body part segments. \cite{reed2016generative} proposed a novel deep architecture and GAN formulation to effectively translating visual concepts from characters to pixels, by adding textual information to both generator and discriminator. They also further investigated the use of additional location, key-points, or segmentation information, to generate images as did by \cite{reed2016learning,reed2016generating}. With only these visual hints as condition and in contrast to our explicit condition on the occluded image, the control exerted over the image generation procedure is still abstract. Many works perform a conditioning over image generation not only on labels or texts but also on images. \cite{zhao2017multi} generated multi-view cloth images from only a single view input by proposing a new image generation model that combines the strengths of the variational inference and the GAN framework. \cite{chen2014inferring} tackled the unseen view inference by casting the problem in terms of tensor completion and adopt a factorization approach to accommodate single-view images. \cite{IsolaPix} provides a general purpose architecture that is effective at synthesizing photos from label maps, reconstructing objects from edge maps, and colorizing images, among other tasks. \cite{yang2015weakly}, \cite{huang2017beyond}, \cite{yim2015rotating}, \cite{ghodrati2015towards} addressed the task of face image generation conditioned on a reference image and a specific face viewpoint. Finally, \cite{yang2017high, yeh2017semantic, pathak2016context, wang2018perceptual} tackled the task of image inpainting where large missing regions have to be filled based on the available visual data. Our work can be seen as a particular case of inpainting, where the portion of the image to inpaint is not known a priori.

% Among the problems of image-to-image translation, inpainting is, with no doubt, the most similar task to ours. 
% The most similar to our task, among the image-to-image translation problems, is, with no doubt, the image inpainting.
\begin{table*}[ht!]
\begin{center}
\caption{Classification performances of our ResNet-101 on RAP dataset}
\label{tab:classfcation_comparison}
\begin{tabular}{l|c|c|c|c|c}
Method & mA & Accuracy & Precision & Recall & F1 \\
\hline\hline
ACN \cite{acn}       		& 69.66 & 62.61 & \bfseries 80.12 & 72.26 & 75.98 \\
DeepMAR \cite{maruno} 	& 73.79 & 62.02 & 74.92 & 76.21 & 75.56 \\
DeepMAR* \cite{rap} 		& 74.44 & 63.67 & 76.53 & 77.47 &  77.00 \\
HP-Net \cite{ydraplus} & 76.12 & 65.39 & 77.33 & 78.79 &  78.05 \\
ACN-Res50 \cite{fabbri2017generative}      & \bfseries 79.73 & 64.13 & 76.96 & 78.72 &  77.83 \\
\hline
Ours      &  78,66 & \bfseries 66,23 & 77.85 & \bfseries 79.71 & \bfseries 78.77\\
\hline
\end{tabular}
\label{tab:classification_comparison}
\end{center}

\end{table*}

\begin{table*}[ht!]
\begin{center}
\caption{Detailed comparison between various pedestrian attribute datasets}
\label{tab:comparison}
\begin{tabular}{l|c|c|c|c|c}
Dataset & \# Scenes & \# Samples & \# Attributes & Min. Resolution & Max. Resolution \\
\hline
\hline
PETA \cite{peta}     & -   & 19,000  & 61(+4) & 17 $\times$ 39  & 169 $\times$ 365 \\
RAP \cite{rap}      & 26  & 41,585  & 69(+3) & 36 $\times$ 92  & 344 $\times$ 554 \\
PA-100K \cite{ydraplus} & 598 & 100,000 & 26     & 50 $\times$ 100 & 758 $\times$ 454 \\
\hline
AiC      & 512 & 125,000 & 24     & 35 $\times$ 85  & 602 $\times$ 1080 \\
\hline
\end{tabular}
\end{center}
\end{table*}
\section{Method}\label{sec:meth}
The goal of our work is to reconstruct occluded body parts of pedestrians in different surveillance scenarios. Given an image of an occluded pedestrian as the network input, we aim at removing the obstructions and replacing them with body parts that could likely belong to the occluded person. Note that, differently from the task of inpainting, we don't want to guide the network with the information about what portion of the image we want to remove and complete. For this purpose, we want to learn an image transformation between pairs of occluded images $I_{OCC}$ and not occluded images $I_{GT}$.

In order to accomplish this, we propose the training procedure depicted in Fig.~\ref{fig:complete_net}: our pipeline takes as input the occluded image $I_{OCC}$, along with the relative attributes $A_{GT}$ and outputs the restored image $I_{GEN}$. $I_{GEN}$ is then inputted to the three networks ResNet-101, VGG-16, and the Discriminator in order to compute the three components of our loss. Each loss component is then backpropagated through the input, updating only the Generator's weights. 

More precisely, to achieve a full body restoration, we train the Generator network $G$ as a feed-forward CNN $G_{\theta_g}$ with parameters $\theta_g$. For $N$ training pairs images $(I_{OCC}, I_{GT})$ and their relative attributes $A_{GT}$, we solve:
\begin{equation}
\label{eq:eg}
    \hat{\theta}_g = \arg \min_{{\theta}_g} \frac{1}{N} \sum_{n=1}^N \mathcal{L}_{total}\left (G_{{\theta}_g}\left (I_{OCC}^n, A_{GT}^n \right), I_{GT}^n\right )
\end{equation}
Here $\hat{\theta}_g$ is obtained by minimizing the loss function $\mathcal{L}_{total}$ described in the next subsection. 
% Differently from what did by \cite{godgan} and  \cite{radford2015unsupervised}, our generator network takes an image as input, along with the attributes, and no random noise vector is used.
% \textcolor{red}{ no random noise vector is used. We know that the task of solve occlusions can be seen as an ill posed problem. In fact, for each non-visible body portion there could be different acceptable reconstructed combinations. In our work we try to solve the problem in a deterministic way, where every single occluded pedestrian should be de-occluded in unique manner(we could say the most acceptable/probable for the implemented network).} 
% instead of a random noise vector, as we want to be deterministic in terms of generated image (e.g. the same person should be always de-occluded in the same manner).
As a result, our generator network learns a mapping from observed images $I_{OCC}$ to output image $I_{GEN}$. This differs from what did by \cite{IsolaPix} and \cite{mirza2014conditional} which use random noise along with the input image.

Following \cite{godgan}, we further define the Discriminator network $D_{\theta_d}$ with parameters $\theta_d$, that we train alongside $G_{\theta_g}$ with the aim of solving the adversarial min-max problem:
\begin{multline}
\label{eq:adv_loss}
    \min_{G} \max_{D} \mathbb{E}_{I_{GT} \sim p_{data}(I_{GT})}[\log{D\left (I_{GT}\right )}] \\
    + \mathbb{E}_{I_{OCC} \sim p_{gen}\left (I_{OCC}, \right)}[\log{1 - D\left (G\left (I_{OCC}, A_{GT}\right)\right)}]
\end{multline}
where $D(I_{GT})$ is the probability of $I_{GT}$ being a ``real'' image while $(1 - D(G(I_{OCC}, A_{GT}))$ is the probability of $G(I_{OCC}, A_{GT})=I_{GEN}$ being a ``fake'' image. The min-max formulation force $G$ to fool the $D$, which is adversarially trained to distinguish between generated ``fake'' images and ``real'' ones. Thanks to this approach, we obtain a Generator network capable of learning solutions that are similar to not occluded images thus indistinguishable by the Discriminator. Note also that, differently from what did by \cite{IsolaPix}, we do not concatenate input images $I_{OCC}$ to the ``fake'' images $I_{GEN}$ or to the ``real'' images $I_{GT}$ as Discriminator input.

\subsection{Loss Function}
The definition of the loss function $\mathcal{L}_{total}$ is crucial for the effectiveness of our Generator network. We propose the following loss formulation, composed by a weighted combination of three components:
\begin{equation}
\label{eq:final_loss}
\mathcal{L}_{total}=\overbrace{\underbrace{\mathcal{L}_{adv}}_\text{adver. loss} + 
\underbrace{\lambda_{1}\cdot\mathcal{L}_{vgg}}_\text{cont. loss} +
\underbrace{\lambda_{2}\cdot\mathcal{L}_{atr}}_\text{attr. loss}}^\text{total loss}
\end{equation}
% \textbf{
% \textbf{ METTERE UNA FIGURA cON UNO SCHEMA COMPLETO E INDICARE DUE DETTAGLI PER LA PARTE ATTRIBUTE
%  volendo citare che in avss abbiamo mostrato that in very simple case a good reconstruction can  obta}in a similar attribute classification. thus here we insert it not as a goal but as a constraint... }

% Going back to equation \ref{eq:adv_loss} we can break it in two parts. The first split concerning the Discriminator D training loss.
% \begin{equation}
% \label{eq:disc_loss}
% \mathcal{L}_{adv}^{D}= \mathcal{L}_{bce}(D(I_{o}), 1)+\mathcal{L}_{bce}(D(G(I_{occ})), 0) 
% \end{equation}
% Where, in the first part of eq. \ref{eq:disc_loss} $D(I_{o})$ indicates the probability that not occluded image is denoted as real by D. While the second component $D(G(I_{occ}))$ means the probability that the generated, not occluded, image is recognized as fake.
The intuition behind this loss formulation is that we want the generated images to contain real people (thanks to $\mathcal{L}_{adv}$), to have similar feature representations (thanks to $\mathcal{L}_{vgg}$) and to preserve visual attributes (thanks to $\mathcal{L}_{atr}$) w.r.t. their non occluded ground truth versions.
\begin{figure*}[h!]
\centering
\includegraphics[width=1\textwidth]{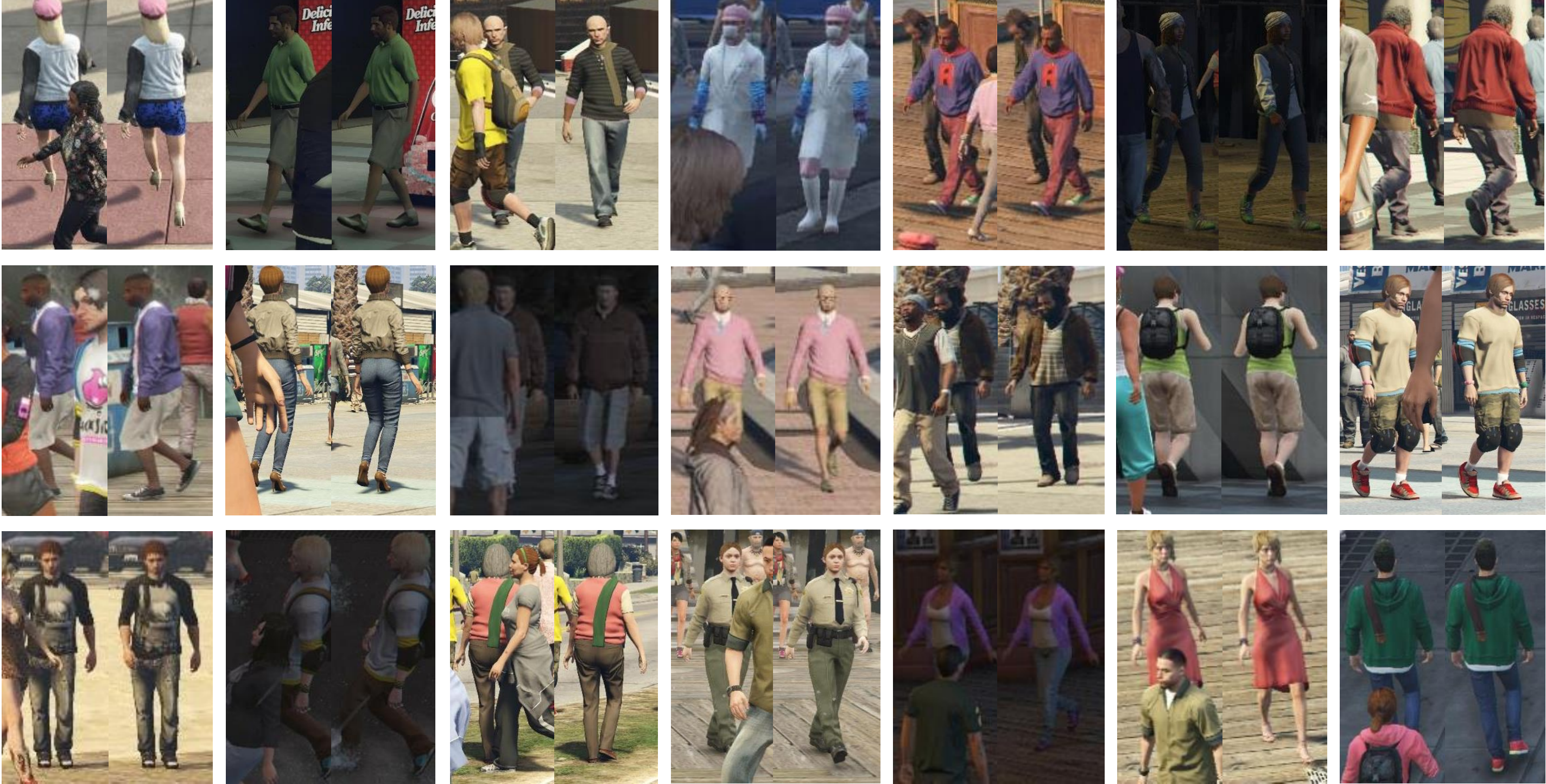}
\caption{Examples from the AiC dataset exhibiting its variety in viewpoints, illuminations and scenarios.}
\label{fig:AiC_images}
\end{figure*}
The first term of Eq.~\eqref{eq:final_loss} is the adversarial loss $\mathcal{L}_{adv}$. This component encourages the Generator network $G$ to generate images belonging to the not occluded domain of pedestrians by fooling the Discriminator network $D$. The adversarial component relative to the Generator network is calculated as follows:
\begin{equation}
\label{eq:gen_loss}
\mathcal{L}_{adv}=-\sum^{N}_{n=1}\log D\left(G\left(I_{OCC}, A_{GT}\right)\right)
\end{equation}
Where $D(G(I_{OCC}, A_{GT}))$ is the probability that $G(I_{OCC}, A_{GT})$ is classified as ``real'' by the discriminator network. Minimizing 
\begin{math}
\log D\left(G\left(I_{OCC}, A_{GT}\right)\right)
\end{math}
instead of
\begin{math}
\log D\left[ 1-\left(G\left(I_{OCC}, A_{GT}\right)\right)\right]
\end{math} is preferred in order to reach a better gradient behavior as indicated by \cite{godgan}.
As a possible drawback, the images produced by the Generator network $G$ are forced to be realistic thanks to the Discriminator network $D$, but they can be unrelated to the original input. For instance, the output could be a plausible image of a pedestrian displaying a very different aspect w.r.t. the input image. Thus, is essential to mix the adversarial loss $\mathcal{L}_{adv}$ with an additional loss, such as L1 or L2, that evaluate the per-pixel distance between the generated and the ground truth image. Usually, training a network using such losses leads to reasonable solutions. However, the outputs appear blurred with lack of high-frequency details.

A possible solution for generating sharper results is to adopt a different content loss, like the perceptual loss introduced by \cite{percloss} and used also by \cite{superes} and \cite{deblurring}:
\begin{multline}
\label{eq:vggloss}
\mathcal{L}_{vgg(i,j)}=\\ \dfrac{1}{W_{i,j}H_{i,j}}\sum_{x=1}^{W_{i,j}}
\sum_{y=1}^{H_{i,j}}\left( \phi_{i,j}(I_{GT})_{x,y} - \phi_{i,j}(I_{GEN})_{x,y}\right)^{2}
\end{multline}

where $W_{i,j}$ and $H_{i,j}$ are the dimensions of the feature maps $\phi_{i,j}$ obtained by the $j$-th convolution after ReLU activation and before the $i$-th max-pooling layer within the VGG16 network, pre-trained on ImageNet by \cite{imagenet_cvpr09}, as done by \cite{percloss}.

The $\mathcal{L}_{vgg}$ that we employed in our work is based on the sum of different intermediate layers of VGG16:
\begin{equation}
\label{eq:ourvggloss}
\mathcal{L}_{vgg}=\sum_{i,j \in L}\mathcal{L}_{vgg(i,j)}
\end{equation}
where $\mathcal{L}_{vgg(i,j)}$ is taken from eq.~\ref{eq:vggloss} and $L$ is the set of used activations. Rather than encouraging the pixels of the output image $I_{GEN}$ to exactly match the pixels of the target image $I_{GT}$, we instead encourage them to have similar feature representations as computed by the VGG16 network. As demonstrated by \cite{percloss} and \cite{mahendran2015understanding}, minimizing the content loss for higher layers do not preserve color and textures. As we reconstruct from early layers, instead, images tend to be perceptually similar to the target image $I_{GT}$ in terms of color and texture.

Since our main purpose is not limited to naively restore the occluded parts of pedestrians, but also to maintain and highlight their attributes, we introduced an additional loss component $\mathcal{L}_{atr}$. Like for the perceptual loss $\mathcal{L}_{vgg}$, we used a classification network as loss function. In particular, we adapted ResNet-101 by \cite{resnet}, pre-trained on ImageNet, to the task of multi-attribute classification. More precisely, we replaced the last two layers (the average pooling and the last fully connected layer) in order to fit the desired input shapes. Differently, from the VGG loss, we work on a higher level of abstraction, forcing the Generator network to produce images that exhibit characteristics coherent with the attributes of the person. In this case, we used a weighted binary cross entropy:
\begin{multline}
\label{eq:attrloss}
	\mathcal{L}_{atr} = -\sum_{i=1}^{N_A} \exp{\left (1-r_i\right )} \cdot\left({y_{i}}\cdot \log\left(\psi_{i}(I_{GEN})\right)\right) \\
	+ \exp{\left (r_i\right )} \cdot \left (1 - y_i \right ) \cdot \log\left (1 - \psi_{i}(I_{GEN} )\right ).
\end{multline}
Here, $N_A$ is the number of attributes classified by the ResNet-101, $r_{i}$ is the positive ratio of $i$-th attribute. $\psi$ is the output of our attribute classification network and $y_{i}$ is the $i$-th ground truth label.

\begin{figure*}[ht!]
\includegraphics[width=1\textwidth]{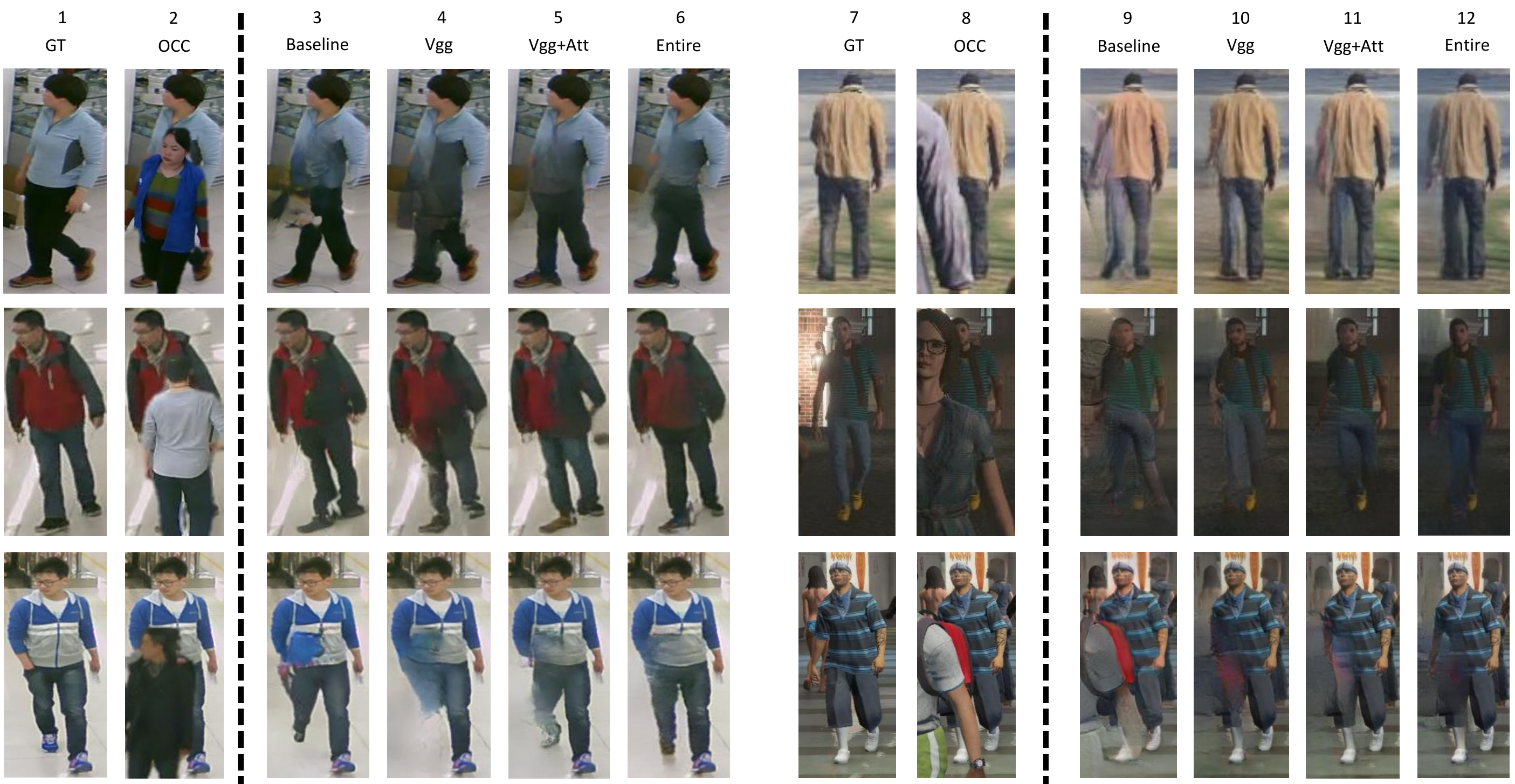}
\caption{Qualitative results based on the ablation study on RAP dataset (leftmost) and AiC dataset (rightmost). GT columns indicate ground truth images while in the OCC columns are presented the input occluded images. Columns 3 and 9 indicate the outputs of our baseline, where adversarial loss and MSE are used. Columns 4 and 10 represents results of the VGG loss. On 5 and 11 we have results of experiments using all the 3 losses combined: adversarial loss, VGG loss, and attribute loss. Finally, columns 6 and 12 show results where attributes are injected as input to the network.}
\label{fig:ablativo}
\end{figure*}

\subsection{Networks Architecture}

\paragraph{Generator Network}
Our Generator structure differs from those presented by \cite{radford2015unsupervised} and \cite{fabbri2017generative}: following \cite{unet} and \cite{IsolaPix} we propose the ``U-Net'' like architecture depicted in Fig.~\ref{fig:genarch}. In particular, the structure of our network slightly differs from the one described by \cite{unet} and \cite{IsolaPix}. The network is composed by 4 down-sampling blocks and a specular number of up-sampling components. Each down-sampling block consists of 2 convolutional layers with a $3 \times 3$ kernel. Each convolutional layer is followed by a batch normalization and a ReLU activation. Finally, each block has a max-pooling layer with stride 2. The up-sampling part has a very similar but overturned structure, where each block is composed by an up-sampling layer of stride 2.  After that, each block is equipped with 2 convolutional layers with a $3\times3$ kernel. The last block has an additional $1\times1$ kernel convolution which is employed to reach the desired number of channels: 3 RGB channels in our case. A $tanh$ has been used as final activation.
We additionally inserted skip connections between mirrored layers, in the down-sampling and up-sampling streams, in order to shuttle low-level information between input and output directly across the network. Eventually, padding is added to avoid cropping the feature maps coming from the skip connections and concatenate them directly to the up-sampling blocks outputs.
Roughly speaking, our task can be seen as a particular case of image-to-image translation, where a mapping is performed between the input image and the output image. Additionally, for the specific problem we are considering, input and output share the same underlying structure despite differing in superficial appearance. Therefore, a rough alignment is present between the two images. In fact, all the non-occluded parts that are visible in the input images must be transferred to the output with no alterations. The structure of the U-Net lends itself optimally to our task, and the skip connections are fundamental for the conservation of the non-occluded image content. In this way, useful low-level information is not lost during the encoding passage: by leveraging this kind of information, we are able to maintain the appearance of visible parts in the image.
\begin{table*}[h]
\begin{center}
\caption{Ablation study results on RAP dataset}
\label{tab:abl_rap_comp}
\begin{tabular}{l|c|c|c|c||c||c|c}
Method & mean Accuracy & Accuracy & Precision & Recall & F1 & SSIM & PSNR\\
\hline\hline
Occlusion       		& 65.74 & 51.06 & 68.72 & 64.36 & 66.47 & 0.7153 & 14.57 \\
\hline
Baseline		& 70.74 & 56.55 & 70.61 & 71.78 & 71.19 & 0.7982 & 20.31 \\
VGG loss		& 72.48 & 58.89 & 72.58 & 73.56 & 73.06 & \bfseries  0.8293 & \bfseries  20.88 \\
VGG and attr. loss & 72.18 & 59.59 & 73.51 & 73.72 & 73.62 & 0.8239 & 20.65 \\
VGG and attr. loss (+ input attr.)		& \bfseries  81.1 & \bfseries  74.8 & \bfseries  84.29 & \bfseries  85.61 & \bfseries  84.94 & 0.8274 & 20.7 \\
\hline
GT data      &  78,66 & 66,23 & 77.85 & 79.71 & 78.77 & - & -\\
\hline
\end{tabular}
\end{center}
\end{table*}

\begin{table*}[h]
\begin{center}
\caption{Ablation study results on AiC dataset}
\label{tab:abl_aic_comp}
\begin{tabular}{l|c|c|c|c||c||c|c}
Method & mean Accuracy & Accuracy & Precision & Recall & F1 & SSIM & PSNR\\
\hline\hline
Occlusion       	& 72.24 & 45.77 & 48.78 & 79.03 & 60.32 & 0.6148 & 18.38 \\
\hline
Baseline		& 72.72 & 45.48 & 48.23 &80.87 & 60.42 & 0.6236 & 20.49 \\
VGG loss		& 78.12 & 53.11 & 55.52 & 85.65  & 67.37 & 0.7088 & 21.5 \\
VGG and attr. loss & 78.37 & 53.3 & 55.73 & 85.46 & 67.46 & \bfseries{0.7101} & \bfseries{21.81} \\
VGG and attr. loss (+ input attr.)		& \bfseries{90.86} & \bfseries{72.15}& \bfseries{74.0} & \bfseries{95.1} & \bfseries{83.23} & 0.6986 & 21.47 \\
\hline
GT data		&  91.89 & 74.87 & 76.80 & 95.43 & 85.11 & - & - \\
\hline
\end{tabular}
\end{center}
\end{table*}
\paragraph{Discriminator Network}
The Discriminator, instead, aims to determine if an image is true or if it has been generated. In particular, the structure is similar to the one proposed by \cite{radford2015unsupervised}, composed by 4 convolutional layers with kernel size $5 \times 5$. The resulting features are followed by one sigmoid activation function in order to obtain a probability for the classification problem. We use batch normalization before every Leaky ReLU activation, except for the first layer.
% \textcolor{red}{\paragraph{ResNet-101} This network operates at a much higher level of abstraction than the normal VGG. In fact, the output return a prediction of the attributes of the reconstructed image. The structure is completely identical to the original ReNet-101. We have only changed the size of the last average pooling layer to accommodate the size of attributes.Finally, instead of the softmax activation, we used a sigmoid. This is because the presence of the attributes is not exclusive, as happens in normal classification problems.}

\subsection{Training Details}
We trained our GAN with $320 \times 128$ resized input images while simultaneously providing the target image in order to compute the supervised losses. We adopted the standard approach by \cite{godgan} to optimize the networks alternating gradient descent updates between the Generator and the Discriminator with $K=1$. Data augmentation is performed by randomly flipping the images horizontally. We used mini-batch SGD applying the Adam solver with momentum parameters $\beta_1 = 0.5$ and $\beta_2 = 0.999$, learning rate $2\cdot{10^{-4}}$ and a batch size of 20. Each training is performed using a Titan Xp GPU.
% In our experiments we try different $\lambda_1$-$\lambda_2$ combination; in section \ref{hpopt} we further explain how we search and choose hyperparameters.

% \subsection{Attribute Classification Network}
% Since we carried out different types of experiments on the occlusion reconstruction network, we needed an objective parameter that would allow us to determine which of the solutions we explored offered the best results.
% To reach this goal we have taken $Resnet-101$, the same introduced in the last subsection,  already pretrained on Imagenet and we did a fine tuning for both our datasets. In addition to this, the only change we made on attribute recongition network was replacing the last two layers in order to have the right shape for our input images.
% The first one is an average pooling layer with a kernel size 10x4, while the last is a fully connected layer dimensionated to have an output vector with the same size of the attributed to be predicted.
% \begin{multline}
% \label{eq:eg}
% 	Loss_{C} = -\sum_{i=1}^A \exp{(1-r_i)} \cdot y_i \cdot \log( \hat{y_i}) \\
% 	+ \exp{(r_i)} \cdot (1 - y_i) \cdot \log(1 - \hat{y_i})
% \end{multline}
% Here the formulation is equal to the one presented before in the \ref{eq:attrloss}. The only difference is that the BCE is not calculated on image generated by G, but on the original images taken from RAP dataset or AiC.   

\section{Datasets}\label{datasets}
We evaluated our method on the RAP dataset, proposed by \cite{rap}, comparing state-of-the-art methods and performing the ablation study over each loss employed. In addition, we further propose a new large-scale computer-graphics dataset AiC for pedestrian attribute recognition in crowded scenes. Differently, from existing publicly available datasets, AiC is mainly focused on occlusion events. % anomaly, phenomenon

\subsection{RAP Dataset} \label{sec:RAPD}
RAP by \cite{rap} is a very richly annotated dataset with 41,585 pedestrian samples, each of which is labeled with 72 attributes as well as viewpoints, occlusions, and body parts information. In order to evaluate our method, we corrupted the dataset with occlusions. Differently, from what did by \cite{fabbri2017generative}, where obstructions were created by cutting parts of images according to regular geometric shapes, we have adopted a more sophisticated approach that has led us to more realistic results.
% \begin{figure}[h!]
% \includegraphics[width=0.5\textwidth]{imgs/mask_finale.jpg}
% \caption{This is a simplified representation of the procedure followed for synthetic occlusion generation. For each image in RAP, a random overlap is computed.}
% \label{fig:maskproc}
% \end{figure}
By exploiting the state-of-the-art performances of Mask R-CNN proposed by \cite{he2017mask}, pre-trained on the COCO Dataset (\cite{lin2014microsoft}), we produced segmentation masks for each person in the RAP dataset. The computed silhouettes were then used to crop people's shapes from the dataset. Those figures are then used to reproduce the occlusions, simply by randomly overlapping the crops to each image sample of RAP dataset. In addition, to reduce the visual gap between the original image and the overlapped person, we performed a Gaussian blurring. However, this is not applied to the whole image but only to the area given by the difference between an expansion and an erosion of the segmentation mask of the overlapping image. The only constraint that we have introduced is that the occluding person must not occupy the portion of the image that has the y coordinate that exceeds the 6/7 of the image height. Each sample is computed as follows:
\begin{equation}
\label{eq:img_occ_formula}
I_{OCC}= I_{GT^1}\odot\neg\alpha\left (\beta\left (I_{GT^2}\right )\right )+\alpha\left (\beta\left (I_{GT^2}\right )\odot I_{GT^2}\right )
\end{equation} 
where $\beta(I_{GT^2})$ is the binary mask generated using Mask R-CNN and morphology operations and $\alpha$ is a function used to translate the overlap section randomly over the destination image $I_{GT^1}$. 
%A simple visual example is presented in figure \ref{fig:maskproc}.
The dataset is already organized in 5 random splits. Each of which contains 33,268 images for training and 8,317 for testing. As did by \cite{rap}, due to the unbalanced distribution of attributes in RAP, we selected the 51 attributes that have the positive example ratio in the dataset higher than 0.01.

\subsection{AiC Dataset}
Most of the publicly available pedestrian attribute datasets, like RAP by \cite{rap}, PETA by \cite{peta} and PA-100K by \cite{ydraplus} does not contemplate occlusion events. They only provide samples of full visible people, completely ignoring crowded situations of pedestrians occluding each other (which is indeed common in urban scenarios). To overcome this limitation, we propose the Attributes in Crowd dataset, a novel synthetic dataset for people attribute recognition in presence of strong occlusions. AiC features 125,000 samples, all being a unique person, each of which is automatically labeled with information concerning sex, age etc. The dataset is split into 100,000 samples for training and 25,000 for testing purposes. Each of the 24 attributes is present at least in a 10\% of samples which highlight a good balance in terms of labels. The collected samples feature a vast number of different body poses, in several urban scenarios with varying illumination conditions and viewpoints. Skeleton joints are also available for each identity. Joints are additionally labeled with an occlusion flag which tells if the specific body part is directly visible from the camera point of view. Moreover, each image sample has his vanilla version where each obstacle is removed from the image. Thus, for each occluded pedestrian, we know exactly how it really is behind the occlusion (this is obviously not achievable in real environments). Fig.~\ref{fig:AiC_images} exhibits some examples of the dataset. AiC was created by exploiting the highly photo-realistic video game \textit{Grand Theft Auto V} developed by \textit{Rockstar North}.
%  For a complete list of attributes, we refer the reader to Tab.~\ref{tab:attributes}.
\begin{figure}[t!]
\centering
\includegraphics{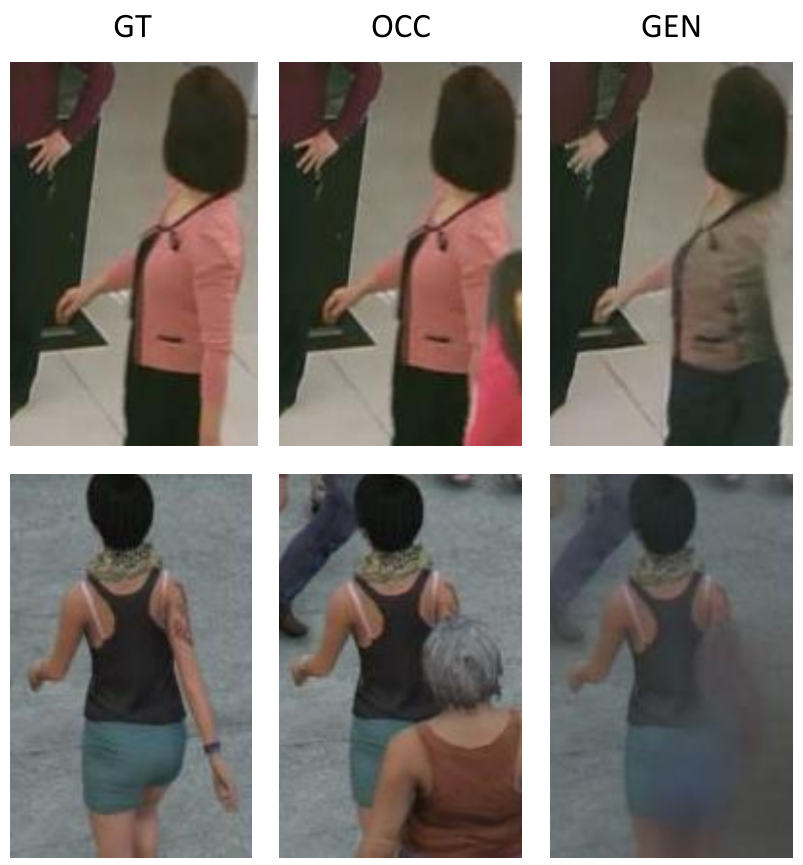} % [width=0.76\linewidth]
\caption{Qualitative results on both RAP and AiC datasets. (first line) an example using a configuration of $\lambda_{1} = 0$ and $\lambda_{2} = 15$ on RAP: the color of the jacket mutate from pink to gray to facilitate the classification, as the majority of jackets in the dataset are dark. (second line) an example using a configuration of $\lambda_{1} = 10$ and $\lambda_{2} = 0.1$ on AiC: blurring the occlusion and not hallucinating new body parts results in a better strategy to facilitate ResNet-101.}
\label{fig:extreme_rap}
\end{figure}

\begin{figure*}[ht!]
\includegraphics[width=1\textwidth]{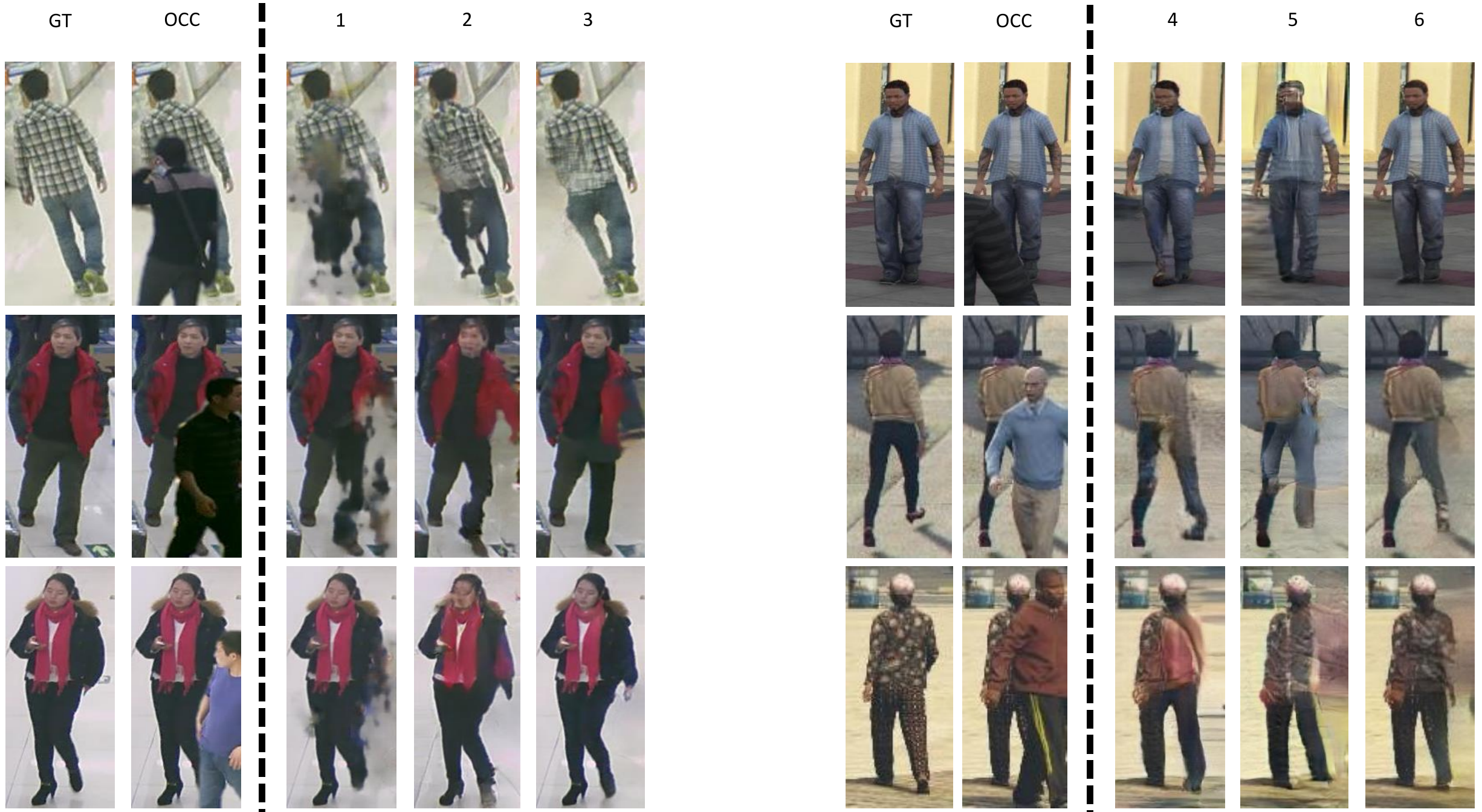}
\caption{Qualitative comparison with state-of-the-art approaches: results are presented for both RAP (leftmost) and AiC (rightmost). GT columns indicate ground truth images while in the OCC columns are presented the input occluded images. Columns 1 and 4 are the images recovered by Pix2Pix by \cite{IsolaPix}. On 2 and 5 are presented results obtained from the method used by \cite{fabbri2017generative}. The last two columns, 3 and 5, show our best comparable approach output (Vgg loss + Attr. loss).}
\label{fig:soacomparison}
\end{figure*}

\section{Experimental Results}
In this section, we provide details about the metrics adopted, followed by a detailed ablation study that presents qualitative and quantitative results for three different combinations of losses (that we added to the adversarial loss): MSE loss, VGG loss and a combination of VGG loss and attribute loss. We also investigate how the information about the attributes of a person can enhance the quality of the produced images. Additionally, we explain the choice of different hyperparameters, exploring their impacts. Finally, we compare our method with the most related works presented by \cite{IsolaPix} and \cite{fabbri2017generative}.

\subsection{Evaluation Metrics} \label{sec:metrics}
Evaluating the quality of synthesized images is an open and challenging problem as stated by \cite{salimans2016improved}. Traditional metrics such as per-pixel MSE do not estimate joint statistics of the result, and therefore do not extrapolate the full structure of the result. In order to more holistically evaluate the visual quality of our results, we employed two tactics. Firstly, we compared the performance of the proposed model through metrics directly calculated over the reconstructed images. Specifically, we adopted the structural similarity SSIM and the peak signal-to-noise ratio PSNR. Secondly, we measured the capability of the proposed network of being able to preserve original attributes, like gender, hairstyle or wearing jacket, by exploiting the ResNet-101 network of \cite{resnet} trained on the task of multi-attribute classification. Thus, following \cite{rap}, \cite{fabbri2017generative} and \cite{ydraplus}, we provide five evaluation metrics for the attribute classification task, namely mean Accuracy, Accuracy, Precision, Recall and F1.

\paragraph{ResNet-101 Classification Network}
We trained the network with $320\times128$ resized images with Adam as optimizer and learning rate set to $2\cdot{10^{-4}}$. In Table \ref{tab:classification_comparison} a comparison on the classification task with other state-of-the-art networks on RAP dataset is presented. The same network is trained independently for each dataset, in order to provide reliable metrics for both RAP and AiC.

\subsection{Ablation Study}
As previously stated, we investigated three loss combinations in order to clarify and highlight the solutions adopted in our work:

\begin{itemize}
\item \textit{Baseline}: the Baseline architecture uses, in conjunction with the adversarial loss, the MSE loss as content loss; 
\item \textit{VGG loss}: differently from the Baseline, we replaced the MSE loss with the VGG loss. The layers (1,2), (2,2), (3,3) and (4,3) are chosen as the set $L$ of activations on Eq. \ref{eq:ourvggloss}. In Eq. \ref{eq:final_loss}, we set $\lambda_{1}$ to 10 and $\lambda_{2}$ to 0 (further details about $\lambda_1$ and $\lambda_{2}$ are presented in the next subsection);
\item \textit{VGG loss + Attr. loss}: in this case, all the three losses are employed. The VGG loss always refers to the same four activation layers. The Attribute loss is computed between the output of the ResNet-101 classification network computed on the generated images and the ground truth labels provided by the datasets. In Eq. \ref{eq:final_loss}, we set $\lambda_{1}$ to 10 and $\lambda_{2}$ to 5 for RAP and to 0.01 for AiC. Note that we did not use all the available attributes of RAP dataset, but only the first 51 for the reason explained at the end of section \ref{sec:RAPD}. For AiC dataset, instead, we used all the available attributes.
\end{itemize}
In order to further investigate how some additional information about the attributes can improve the restoration process, we performed a further experiment where attributes are fed as input to the network, along with the occluded image:
\begin{itemize}
\item \textit{Entire}: in this setup, we adopted both the VGG loss and the Attribute loss, along with the adversarial loss. Differently, from our main method, attributes are injected directly to the main flow of the Generator network. Specifically, the attribute vector of the occluded pedestrian is fed to a fully connected layer in order to produce a feature vector that is reshaped to match the bottleneck dimension of our Generator network. 
\end{itemize}
Fig.~\ref{fig:ablativo} shows some qualitative results. The baseline performs considerably worse than the other setups, not being able to completely remove the occlusions on AiC (column 9 of Fig.~\ref{fig:ablativo}). This is probably due to the fact that AiC is a more challenging dataset compared to our corrupted version of RAP. For the same reason, RAP results are overall more appealing than the ones of AiC. Moreover, no substantial difference appears between the other setups, highlighting the fact that the VGG loss is the main component that guides the network to produce high-quality results. 
\begin{table*}[h!]
\begin{center}
\caption{Comparison with the state-of-the-art method on RAP dataset}
\label{tab:art_rap_comp}
\begin{tabular}{l|c|c|c|c||c||c|c}
Method & mA & Accuracy & Precision & Recall & F1 & SSIM & PSNR\\
\hline\hline
Occlusion       		& 65.74 & 51.06 & 68.72 & 64.36 & 66.47 & 0.7153 & 14.57 \\
\hline
Pix2Pix \cite{IsolaPix}		& 69.49 & 52.05 & 65.07 & 70.06 & 67.47 & 0.7348 & 17.91 \\
\cite{fabbri2017generative}  & 65.92 & 51.44 & 65.77 & 67.94 & 66.84 & 0.6798 & 18.4 \\
Ours & \bfseries 72.18 & \bfseries 59.59 & \bfseries 73.51 & \bfseries 73.72 & \bfseries 73.62 & \bfseries 0.8239 & \bfseries 20.65 \\
\hline
\end{tabular}
\end{center}
\end{table*}
\begin{table*}[h]
\begin{center}
\caption{Comparison with the state-of-the-art method on AiC dataset}
\label{tab:art_aic_comp}
\begin{tabular}{l|c|c|c|c||c||c|c}
Method & mA & Accuracy & Precision & Recall & F1 & SSIM & PSNR\\
\hline\hline
Occlusion       		& 72.24 & 45.77 & 48.78 & 79.03 & 60.32 & 0.6148 & 18.38 \\
\hline
Pix2Pix \cite{IsolaPix}			& 71.93 & 44.27 & 46.75 & 81.61 & 59.45 & 0.6351 & 21.22 \\
\cite{fabbri2017generative}  & 67.14 & 38.21 & 40.61 & 79.9 & 53.85 & 0.573 & 20.11 \\
Ours 	& \bfseries  78.37 & \bfseries  53.3 & \bfseries  55.73 & \bfseries  85.46 & \bfseries  67.46 & \bfseries 0.7101 & \bfseries 21.81 \\
\hline
\end{tabular}
\end{center}
\end{table*}
Table \ref{tab:abl_rap_comp} and Table \ref{tab:abl_aic_comp} present quantitative results for RAP and AiC respectively based on our ablation study. The tables also provide metrics referred to the occluded images before the restoration process. By observing the tables, we can state that, despite being visually indistinguishable, the images obtained from the VGG loss and from our Entire configuration produce very different results in terms of attribute metrics. We can also observe that there is no substantial difference between the VGG loss and the VGG loss with Attributes loss. In fact, RAP shows a gap of one percentage point in almost all the classification metrics, while AiC shows very little differences, probably due to the more challenging nature of AiC. Moreover, Table \ref{tab:abl_rap_comp} shows that the Entire setup reach higher scores compared to the upper bound of the ground truth images. Also Table \ref{tab:abl_aic_comp} shows performances that are close to the ground truth metrics when we input attribute information directly to the Generator. In fact, with attributes as input, the Generator network, by restoring the occluded images, is able to produce an output that has enhanced attribute characteristics (although this is not visible to the naked eye). As can be shown in the next subsection, further forcing the generation output on classification metrics, we can reach results that exceed the ground truth upper bound even on AiC, at a price of a drop on reconstruction metrics. 

\subsection{Hyperparameter Optimization}\label{hpopt}
Hyperparameter tuning is a crucial aspect in designing machine learning frameworks, as the performance of an algorithm can be highly dependent on the choice of hyperparameters. In fact,  $\lambda_1$ and $\lambda_2$ were selected using a grid search technique. In particular, we searched for a trade off between classification metrics (accuracy, precision, recall, f1) and pixel-level reconstruction metrics (PSNR, SSIM). We performed a different grid search for four different configurations combining each dataset with the two main setups: VGG loss + Attr. loss and the Entire pipeline. 

For what concerns the VGG loss + Attr. loss setup, we observed that, in general, a configuration with $\lambda_{1} \gg \lambda_{2}$ brings to better pixel-level reconstruction metrics but poor classification performances. On the other hand, solutions with $\lambda_{1} \ll \lambda_{2}$ show good classification performances but low pixel-level reconstruction metrics. Also, increasing $\lambda_1$ over the value of 10 does not further improve PSNR and SSIM metrics (for both RAP and AiC). The same behavior happens for $\lambda_2$: the classification metrics do not improve for values greater than 5 (for RAP) and 0.01 (for AiC). This difference of $\lambda_2$ between the two datasets may be caused by the fact that AiC is more challenging than RAP. In fact, during training, the Attributes Loss on AiC is orders of magnitude greater than the same loss on RAP, thus, a smaller $\lambda_2$ is needed to maintain the balance between the losses.

For what concern the Entire pipeline, we observed a different behavior on $\lambda_2$: increasing $\lambda_2$ does steadily improve the classification metrics (reaching up to 98.89 mean Accuracy with $\lambda_2 = 5$) while drastically decreasing PSNR and SSIM. This behavior happens on both RAP and AiC. By giving more importance to the Attributes Loss, the Generator network is able to enhance attribute characteristics to the point that they are highly recognizable by the classification network, at the price of not maintaining low-level similarity. Fig. \ref{fig:extreme_rap} shows a direct consequence at qualitative level on both RAP and AiC. The first line depicts an extreme configuration of $\lambda_{1} = 0$ and $\lambda_{2} = 15$ on RAP. With no low-level constraints, the Generator network is able to mutate the color of the jacket to facilitate the ResNet-101 ``jacket attribute'' recognition. The second line of Fig.~\ref{fig:extreme_rap}, instead, shows an example obtained using $\lambda_{1} = 10$ and $\lambda_{2} = 0.1$ on AiC. In this case, the behavior is completely different: due to the high diversity of attributes in AiC, the Generator learns to simply remove the obstacle, not adding (hallucinating) many details to the removed portion of the image. Adding imprecise details would, in fact, mislead the attribute classification network.

\subsection{Comparison Against State-Of-The-Art Techniques}
Since our task of de-occlusion is novel, there are no direct works to compare with. So, to match the results of our network, in addition to our previous work, we also retrained the Pix2Pix framework on both RAP and AiC.
\paragraph{\textbf{Our Previous work}}
Like our current method, \cite{fabbri2017generative} exploits an adversarial based framework to achieve a translation from an occluded-pedestrian domain to a completely visible body domain. The main difference with our current method resides in the loss formulation: \cite{fabbri2017generative} minimizes a combination of adversarial loss and sum of squared error loss (SSE), completely ignoring high-level and low-level similarities. Another important difference lies in the Generator architecture: our previous work uses a simple hourglass architecture with no \textit{skip connections}, while in our current method we adopted a U-net based solution. The U-net architecture shows better performances in tasks where some input information has to be shuttled directly to the output with no variation. In fact, as can be seen in Fig. \ref{fig:soacomparison}, our previous work fails to preserve the portions of the image that should remain unchanged (especially the faces). 
\paragraph{\textbf{Pix2Pix}} 
% For a fair comparison we use the public code available on github. We must specify that the complete project was thought to work with square size image, in particular: $128\times128$ and $256\times256$. So, to use it with our images($320\times128$) we took only the network basic blocks.
\cite{IsolaPix} investigates conditional adversarial networks as a general-purpose solution to image-to-image translation problems. As in our Generator network, Pix2Pix exploits a U-net based architecture. The only substantial architectural difference is in the number of convolutional layers before each downsampling and after each upsampling operation. Also, the Discriminators differs: Pix2Pix uses a patch level discriminator that only penalizes structure at the scale of patches, while in our work we adopt an image level discriminator that takes the whole image as input. A patch level discriminator models the image as a Markov random field, assuming independence between pixels separated by more than a patch diameter. This is indeed not the case when dealing with images of people. In fact, for example, the skin color of the face should match the skin color of the hands. Also, the trousers are usually made of the same color. Another significant difference lies in the content loss: Pix2Pix, like our previous work, uses a pixel-level loss (L1 instead of SSE), assuming pixel independence, and forcing pixels of the output image to exactly match the pixels of the target image. In our work, instead, we exploit a combination of high-level and low-level consistency by encouraging the overall images the have similar feature representations as computed by the VGG16 network, and similar visual attributes as computed by the ResNet-101.

From Table \ref{tab:art_rap_comp} and Table \ref{tab:art_aic_comp} it can be shown that our network perform favourably for each metric, both for RAP and AiC datasets. From Fig. \ref{fig:soacomparison} it also emerges that our method, despite not using attention mechanisms, is able to detect and to remove the occlusion, with no external additional information. Furthermore, differently from the works by\cite{fabbri2017generative} and \cite{IsolaPix}, our method learns to transfer with no alterations the portion of images that are not occluded. 
% Finally, Fig. \ref{fig:failure} depicts some failure cases of our method that display the challenge of strong occlusions.

%\begin{figure}[ht!]
%\centering
%\includegraphics[width=0.45\textwidth]{imgs/new_error.pdf}
%\caption{Some failure cases on RAP (leftmost) and AiC (rightmost). From this images, it is possible to see that, in cases of strong occlusions, complete restoration remains a difficult task. Moreover, also with particular background conditions, as we can see in the first image triplet, the network is not able to perform a reliable reconstruction.}
%\label{fig:failure}
%\end{figure}

% \begin{table*}[htb]
% \begin{center}
% \caption{AiC attribute details}
% \begin{tabular}{l|c}
% Attribute & Ratio\\
% \hline\hline
% Female		& 53.64 \\
% Age 17-30	& 88.00 \\
% Age 31-45	& 12.00 \\
% BodyNormal	& 33.44 \\
% BodyThin	& 66.56 \\
% BaldHead	& 6.85 \\
% LongHair	& 72.62 \\
% BlackHair	& 33.57\\
% Hat			& 8.35\\
% Muffler		& 6.93\\
% Shirt		& 17.78\\
% Sweater		& 0.41\\
% Jacket		& 20.35\\
% TightHood	& 18.23\\
% Dress		& 8.87\\
% ShortSleeve	& 19.87\\
% LongTrousers	& 28.84\\
% Skirt	& 9.49\\
% Jeans	& 24.75\\
% Tights	& 6.55\\
% shoesLeather	& 11.65\\
% shoesSport	& 33.32\\
% shoesBoots	& 11.78\\
% Backpack & 3.48\\

% \end{tabular}
% \end{center}
% \label{tab:attributes}
% \end{table*}

\section{Conclusions}
In this work, we presented the use of GANs for image enhancing in people attributes classification. Our generator network has been designed to overcome a common problem in surveillance scenarios, namely people occlusion. Experiments have shown that jointly enhancing images before feeding them to an attribute classification network can improve the results even when input images are affected by this issue. We think that this line of work can foster research about the problem of attribute classification in surveillance contexts, where camera resolution and positioning cannot be neglected.
% In the worst
% case scenario, even though the replaced body parts does not
% reflect the real attributes of the person, the reconstruction
% still helps the classifier: by removing the occlusion we produce
% an image that contains only the subject without noise
% that could lead to misclassifications. For example, an image
% containing a person occluded by another person could
% induce the network to classify the attributes of the person in
% the foreground which is not the subject of the image.

\section*{Acknowledgments}
The work is supported by the Italian MIUR, Ministry of Education, Universities and Research, under the project COSMOS PRIN 2015 programme 201548C5NT. We also gratefully acknowledge the support of Panasonic Silicon Valley Lab and Facebook AI Research with the donation of the GPUs used for this research.

% \section*{Acknowledgments}
% Acknowledgments should be inserted at the end of the paper, before the
% references, not as a footnote to the title. Use the unnumbered 
% Acknowledgements Head style for the Acknowledgments heading.

% \bibliographystyle{model2-names}
% \bibliography{egbib}

{\small
\bibliographystyle{ieee}
\bibliography{egbib}

\begin{thebibliography}{10}\itemsep=-1pt

\bibitem{chen2014inferring}
C.-Y. Chen and K.~Grauman.
\newblock Inferring unseen views of people.
\newblock In {\em Proceedings of the IEEE conference on computer vision and
  pattern recognition}, pages 2003--2010, 2014.

\bibitem{track3}
D.~Coppi, S.~Calderara, and R.~Cucchiara.
\newblock Transductive people tracking in unconstrained surveillance.
\newblock {\em IEEE Transactions on Circuits and Systems for Video Technology},
  26(4):762--775, 2016.

\bibitem{imagenet_cvpr09}
J.~Deng, W.~Dong, R.~Socher, L.-J. Li, K.~Li, and L.~Fei-Fei.
\newblock {ImageNet: A Large-Scale Hierarchical Image Database}.
\newblock In {\em CVPR09}, 2009.

\bibitem{peta}
Y.~Deng, P.~Luo, C.~C. Loy, and X.~Tang.
\newblock Pedestrian attribute recognition at far distance.
\newblock In {\em Proceedings of the 22Nd ACM International Conference on
  Multimedia}, 2014.

\bibitem{fabbri2017generative}
M.~Fabbri, S.~Calderara, and R.~Cucchiara.
\newblock Generative adversarial models for people attribute recognition in
  surveillance.
\newblock In {\em Advanced Video and Signal Based Surveillance (AVSS), IEEE
  International Conference on}. IEEE, 2017.

\bibitem{fabbri2018learning}
M.~Fabbri, F.~Lanzi, S.~Calderara, A.~Palazzi, R.~Vezzani, and R.~Cucchiara.
\newblock Learning to detect and track visible and occluded body joints in a
  virtual world.
\newblock {\em arXiv preprint arXiv:1803.08319}, 2018.

\bibitem{ghodrati2015towards}
A.~Ghodrati, X.~Jia, M.~Pedersoli, and T.~Tuytelaars.
\newblock Towards automatic image editing: Learning to see another you.
\newblock {\em arXiv preprint arXiv:1511.08446}, 2015.

\bibitem{godgan}
I.~Goodfellow, J.~Pouget-Abadie, M.~Mirza, B.~Xu, D.~Warde-Farley, S.~Ozair,
  A.~Courville, and Y.~Bengio.
\newblock Generative adversarial nets.
\newblock In {\em Advances in Neural Information Processing Systems 27}, pages
  2672--2680. Curran Associates, Inc., 2014.

\bibitem{he2017mask}
K.~He, G.~Gkioxari, P.~Doll{\'a}r, and R.~Girshick.
\newblock Mask r-cnn.
\newblock In {\em Computer Vision (ICCV), 2017 IEEE International Conference
  on}, pages 2980--2988. IEEE, 2017.

\bibitem{resnet}
K.~He, X.~Zhang, S.~Ren, and J.~Sun.
\newblock Deep residual learning for image recognition.
\newblock In {\em {CVPR}}, pages 770--778. {IEEE} Computer Society, 2016.

\bibitem{action}
S.~Herath, M.~Harandi, and F.~Porikli.
\newblock Going deeper into action recognition: A survey.
\newblock {\em Image and Vision Computing}, 60:4 -- 21, 2017.
\newblock Regularization Techniques for High-Dimensional Data Analysis.

\bibitem{huang2017beyond}
R.~Huang, S.~Zhang, T.~Li, R.~He, et~al.
\newblock Beyond face rotation: Global and local perception gan for
  photorealistic and identity preserving frontal view synthesis.
\newblock {\em arXiv preprint arXiv:1704.04086}, 2017.

\bibitem{IsolaPix}
P.~Isola, J.~Zhu, T.~Zhou, and A.~A. Efros.
\newblock Image-to-image translation with conditional adversarial networks.
\newblock In {\em 2017 {IEEE} Conference on Computer Vision and Pattern
  Recognition, {CVPR} 2017, Honolulu, HI, USA, July 21-26, 2017}, pages
  5967--5976, 2017.

\bibitem{percloss}
J.~Johnson, A.~Alahi, and L.~Fei{-}Fei.
\newblock Perceptual losses for real-time style transfer and super-resolution.
\newblock In {\em Computer Vision - {ECCV} 2016 - 14th European Conference,
  Amsterdam, The Netherlands, October 11-14, 2016, Proceedings, Part {II}},
  pages 694--711, 2016.

\bibitem{kingma2013auto}
D.~P. Kingma and M.~Welling.
\newblock Auto-encoding variational bayes.
\newblock {\em arXiv preprint arXiv:1312.6114}, 2013.

\bibitem{deblurring}
O.~Kupyn, V.~Budzan, M.~Mykhailych, D.~Mishkin, and J.~Matas.
\newblock Deblurgan: Blind motion deblurring using conditional adversarial
  networks.
\newblock {\em CoRR}, abs/1711.07064, 2017.

\bibitem{lassner2017generative}
C.~Lassner, G.~Pons-Moll, and P.~V. Gehler.
\newblock A generative model of people in clothing.
\newblock In {\em Proceedings of the IEEE International Conference on Computer
  Vision}, volume~6, 2017.

\bibitem{superes}
C.~Ledig, L.~Theis, F.~Huszar, J.~Caballero, A.~Cunningham, A.~Acosta, A.~P.
  Aitken, A.~Tejani, J.~Totz, Z.~Wang, and W.~Shi.
\newblock Photo-realistic single image super-resolution using a generative
  adversarial network.
\newblock In {\em 2017 {IEEE} Conference on Computer Vision and Pattern
  Recognition, {CVPR} 2017, Honolulu, HI, USA, July 21-26, 2017}, pages
  105--114, 2017.

\bibitem{maruno}
D.~Li, X.~Chen, and K.~Huang.
\newblock Multi-attribute learning for pedestrian attribute recognition in
  surveillance scenarios.
\newblock {\em 2015 3rd IAPR Asian Conference on Pattern Recognition (ACPR)},
  pages 111--115, 2015.

\bibitem{rap}
D.~Li, Z.~Zhang, X.~Chen, H.~Ling, and K.~Huang.
\newblock A richly annotated dataset for pedestrian attribute recognition.
\newblock {\em preprint arXiv:1603.07054}, 2016.

\bibitem{lin2014microsoft}
T.-Y. Lin, M.~Maire, S.~Belongie, J.~Hays, P.~Perona, D.~Ramanan,
  P.~Doll{\'a}r, and C.~L. Zitnick.
\newblock Microsoft coco: Common objects in context.
\newblock In {\em European conference on computer vision}, pages 740--755.
  Springer, 2014.

\bibitem{ydraplus}
X.~Liu, H.~Zhao, M.~Tian, L.~Sheng, J.~Shao, J.~Yan, and X.~Wang.
\newblock Hydraplus-net: Attentive deep features for pedestrian analysis.
\newblock In {\em Proceedings of the IEEE international conference on computer
  vision}, pages 1--9, 2017.

\bibitem{reid}
X.~Ma, X.~Zhu, S.~Gong, X.~Xie, J.~Hu, K.-M. Lam, and Y.~Zhong.
\newblock Person re-identification by unsupervised video matching.
\newblock {\em Pattern Recognition}, 65:197 -- 210, 2017.

\bibitem{mahendran2015understanding}
A.~Mahendran and A.~Vedaldi.
\newblock Understanding deep image representations by inverting them.
\newblock In {\em Proceedings of the IEEE conference on computer vision and
  pattern recognition}, pages 5188--5196, 2015.

\bibitem{mirza2014conditional}
M.~Mirza and S.~Osindero.
\newblock Conditional generative adversarial nets.
\newblock {\em arXiv preprint arXiv:1411.1784}, 2014.

\bibitem{op2015detection}
R.~M. op~het Veld, R.~Wijnhoven, Y.~Bondarev, et~al.
\newblock Detection and handling of occlusion in an object detection system.
\newblock In {\em Video Surveillance and Transportation Imaging Applications
  2015}, volume 9407, page 94070N. International Society for Optics and
  Photonics, 2015.

\bibitem{ouyang2016partial}
W.~Ouyang, X.~Zeng, and X.~Wang.
\newblock Partial occlusion handling in pedestrian detection with a deep model.
\newblock {\em IEEE Transactions on Circuits and Systems for Video Technology},
  26(11):2123--2137, 2016.

\bibitem{track1}
J.~Pan and B.~Hu.
\newblock Robust occlusion handling in object tracking.
\newblock In {\em 2007 IEEE Conference on Computer Vision and Pattern
  Recognition}, pages 1--8, 2007.

\bibitem{pathak2016context}
D.~Pathak, P.~Krahenbuhl, J.~Donahue, T.~Darrell, and A.~A. Efros.
\newblock Context encoders: Feature learning by inpainting.
\newblock In {\em Proceedings of the IEEE Conference on Computer Vision and
  Pattern Recognition}, pages 2536--2544, 2016.

\bibitem{radford2015unsupervised}
A.~Radford, L.~Metz, and S.~Chintala.
\newblock Unsupervised representation learning with deep convolutional
  generative adversarial networks.
\newblock {\em arXiv preprint arXiv:1511.06434}, 2015.

\bibitem{reed2016generative}
S.~Reed, Z.~Akata, X.~Yan, L.~Logeswaran, B.~Schiele, and H.~Lee.
\newblock Generative adversarial text to image synthesis.
\newblock {\em arXiv preprint arXiv:1605.05396}, 2016.

\bibitem{reed2016generating}
S.~Reed, A.~van~den Oord, N.~Kalchbrenner, V.~Bapst, M.~Botvinick, and
  N.~de~Freitas.
\newblock Generating interpretable images with controllable structure.
\newblock 2016.

\bibitem{reed2016learning}
S.~E. Reed, Z.~Akata, S.~Mohan, S.~Tenka, B.~Schiele, and H.~Lee.
\newblock Learning what and where to draw.
\newblock In {\em Advances in Neural Information Processing Systems}, pages
  217--225, 2016.

\bibitem{rezende2014stochastic}
D.~J. Rezende, S.~Mohamed, and D.~Wierstra.
\newblock Stochastic backpropagation and approximate inference in deep
  generative models.
\newblock {\em arXiv preprint arXiv:1401.4082}, 2014.

\bibitem{Guler2018DensePose}
I.~K. Riza Alp~Guler, Natalia~Neverova.
\newblock Densepose: Dense human pose estimation in the wild.
\newblock 2018.

\bibitem{unet}
O.~Ronneberger, P.Fischer, and T.~Brox.
\newblock U-net: Convolutional networks for biomedical image segmentation.
\newblock In {\em Medical Image Computing and Computer-Assisted Intervention
  (MICCAI)}, volume 351 of {\em LNCS}, pages 234--241. Springer, 2015.

\bibitem{salimans2016improved}
T.~Salimans, I.~Goodfellow, W.~Zaremba, V.~Cheung, A.~Radford, and X.~Chen.
\newblock Improved techniques for training gans.
\newblock In {\em Advances in Neural Information Processing Systems}, pages
  2234--2242, 2016.

\bibitem{reid3}
A.~Subramaniam, M.~Chatterjee, and A.~Mittal.
\newblock Deep neural networks with inexact matching for person
  re-identification.
\newblock In D.~D. Lee, M.~Sugiyama, U.~V. Luxburg, I.~Guyon, and R.~Garnett,
  editors, {\em Advances in Neural Information Processing Systems 29}, pages
  2667--2675. Curran Associates, Inc., 2016.

\bibitem{acn}
P.~Sudowe, H.~Spitzer, and B.~Leibe.
\newblock Person attribute recognition with a jointly-trained holistic cnn
  model.
\newblock In {\em ICCV Workshops}, pages 329--337. IEEE Computer Society, 2015.

\bibitem{wang2018perceptual}
C.~Wang, C.~Xu, C.~Wang, and D.~Tao.
\newblock Perceptual adversarial networks for image-to-image transformation.
\newblock {\em IEEE Transactions on Image Processing}, 27(8):4066--4079, 2018.

\bibitem{track2}
X.~Wang, Z.~Hou, W.~Yu, L.~Pu, Z.~Jin, and X.~Qin.
\newblock Robust occlusion-aware part-based visual tracking with object scale
  adaptation.
\newblock {\em Pattern Recognition}, 81:456 -- 470, 2018.

\bibitem{yan2016attribute2image}
X.~Yan, J.~Yang, K.~Sohn, and H.~Lee.
\newblock Attribute2image: Conditional image generation from visual attributes.
\newblock In {\em European Conference on Computer Vision}, pages 776--791.
  Springer, 2016.

\bibitem{yang2017high}
C.~Yang, X.~Lu, Z.~Lin, E.~Shechtman, O.~Wang, and H.~Li.
\newblock High-resolution image inpainting using multi-scale neural patch
  synthesis.
\newblock In {\em The IEEE Conference on Computer Vision and Pattern
  Recognition (CVPR)}, volume~1, page~3, 2017.

\bibitem{yang2015weakly}
J.~Yang, S.~E. Reed, M.-H. Yang, and H.~Lee.
\newblock Weakly-supervised disentangling with recurrent transformations for 3d
  view synthesis.
\newblock In {\em Advances in Neural Information Processing Systems}, pages
  1099--1107, 2015.

\bibitem{yeh2017semantic}
R.~A. Yeh, C.~Chen, T.~Y. Lim, A.~G. Schwing, M.~Hasegawa-Johnson, and M.~N.
  Do.
\newblock Semantic image inpainting with deep generative models.
\newblock In {\em Proceedings of the IEEE Conference on Computer Vision and
  Pattern Recognition}, pages 5485--5493, 2017.

\bibitem{yim2015rotating}
J.~Yim, H.~Jung, B.~Yoo, C.~Choi, D.~Park, and J.~Kim.
\newblock Rotating your face using multi-task deep neural network.
\newblock In {\em Proceedings of the IEEE Conference on Computer Vision and
  Pattern Recognition}, pages 676--684, 2015.

\bibitem{zhao2017multi}
B.~Zhao, X.~Wu, Z.-Q. Cheng, H.~Liu, Z.~Jie, and J.~Feng.
\newblock Multi-view image generation from a single-view.
\newblock {\em arXiv preprint arXiv:1704.04886}, 2017.

\bibitem{reid1}
J.~Zhuo, Z.~Chen, J.~Lai, and G.~Wang.
\newblock Occluded person re-identification.
\newblock {\em CoRR}, abs/1804.02792, 2018.

\end{thebibliography}
}

\end{document}